\documentclass{article}
\usepackage{parskip}
\usepackage[margin=1.0in]{geometry}

\usepackage{graphicx}
\usepackage{caption}
\usepackage{subcaption}
\usepackage{tikz}
\usepackage{float}

\usepackage{physics}
\usepackage{amsmath}
\usepackage{amsthm}
\usepackage{amstext}
\usepackage{amssymb}
\usepackage{thmtools}
\usepackage{siunitx}
\usepackage{bm}

\theoremstyle{plain}

\theoremstyle{definition}

\theoremstyle{remark}

\usepackage[english]{babel}

\usepackage{pdfpages}

\usepackage{dsfont}
\usepackage{xcolor}
\definecolor{amber}{rgb}{1.0, 0.6, 0.0}
\usepackage{textcomp}
\usepackage{listings}
\usepackage{relsize}
\usepackage{multicol}
\usepackage{hyperref}
\hypersetup{
	colorlinks=true,
	linkcolor=black,
	filecolor=black,      
	urlcolor=black,
	citecolor=blue,
}
\usepackage[capitalize,noabbrev]{cleveref}
\crefname{figure}{Fig.}{Fig.}
\crefname{table}{Table}{Tables}
\crefname{equation}{Eqn.}{Eqns.}
\crefname{algocf}{Algorithm}{Algorithms}
\crefname{exmp}{Example}{Ex.}

\crefname{lemma}{Lemma}{Lemmas}
\crefname{corollary}{Corollary}{Cor.}
\crefname{defn}{Definition}{Definitions}

\usepackage{physics}
\usepackage{url}

\usepackage{pifont}

\usepackage{bbm}
\usepackage{mathtools}
\usepackage{hhline}
\usepackage{booktabs}
\usepackage{multirow}

\newcommand{\R}{\mathbb{R}}

\newcommand{\mcE}{\mathcal{E}}

\newcommand{\pinnNet}{\mathcal{N}}
\newcommand{\uFlattenedLetter}{u}

\usepackage[ruled,vlined]{algorithm2e}

\usepackage{authblk}

\title{Data-Driven, ML-assisted Approaches to Problem Well-Posedness}

\author[1,*]{Tom Bertalan}
\author[2,6]{George A. Kevrekidis}
\author[3]{Eleni D. Koronaki}
\author[4]{Siddhartha Mishra}
\author[5]{Elizaveta Rebrova}
\author[2]{Yannis G. Kevrekidis}

\affil[1]{Amgen, California, USA}
\affil[2]{Department of Applied Mathematics and Statistics, Johns Hopkins University, Baltimore, MD, USA}
\affil[3]{Faculty of Science, Technology and Medicine, University of Luxembourg, Esch-sur-Alzette, Luxembourg}
\affil[4]{Seminar for Applied Mathematics, Department of Mathematics, and ETH AI Center, ETH Zurich, Zurich, Switzerland}
\affil[5]{Department of Operations Research and Financial Engineering, Princeton University, Princeton, NJ, USA }
\affil[6]{Los Alamos National Laboratory, Los Alamos, NM, USA}
\affil[*]{This work was performed while T. Bertalan was a post-doctoral researcher at Department of Chemical and Biomolecular Engineering, Johns Hopkins University, Baltimore, MD, USA}

\date{March 2025\\LA-UR-25-22386}

\begin{document}

\maketitle

\begin{abstract}
Classically, to solve differential equation problems, it is necessary to specify sufficient initial and/or boundary conditions so as to allow the existence of a unique solution. Well-posedness of differential equation problems thus involves studying the existence and  uniqueness of solutions, and their dependence to such pre-specified conditions. However, in part due to mathematical necessity, these conditions are usually specified “to arbitrary precision” only on (appropriate portions of) the boundary of the space-time domain. This does not mirror how data acquisition is performed in realistic situations, where one may observe entire “patches” of solution data at arbitrary space-time locations; alternatively one might have access to more than one solutions stemming from the same differential operator. In our short work, we demonstrate how standard tools from machine and manifold learning can be used to infer, in a data driven manner, certain well-posedness features of differential equation problems,  for initial/boundary condition combinations under which rigorous existence/uniqueness theorems are not known. Our study naturally combines a data assimilation perspective with an operator-learning one.
\end{abstract}

\section{Introduction}

Take an operator (a simple, linear one, like $u_{xx}=0$); discretize it, say with finite differences, into a set of linear equations; and then solve these equations between, say, $x=0$ and $x=1$ -- but (a) do it variationally, with a random seed for an off-the-shelf nonlinear optimizer, and (b) do it many times (as we will see below, what we do has intimate connections to randomized numerical linear algebra). The outcome is more or less easy to guess: if one prescribes \textit{two conditions} --whether initial ($u(0), u_x(0)$), or boundary ($u(0), u(1)$), or interior (say $u(0.2), u(0.9)$)-- there will be a unique (discretized) solution (with zero residuals, to machine precision). If one prescribes a single condition, there will be \textit{a one-parameter infinite family of solutions} (zero residuals to machine precision). If one prescribes zero boundary conditions, there will be \textit{a two-parameter infinity of solutions} (zero residuals). But if one prescribes three conditions, what happens? If they are compatible, a unique solution (zero residuals). If incompatible, no solutions --- but what will the variational solver converge to? 

Clearly, statistical processing of sets of randomized solutions (more accurately, randomized outcomes of the variational solution process) for this simple problem can provide information about its well-posedness: whether the solution exists, and whether it is unique. And when neither of these things is the case, the reader may already perceive that studying the statistics of the results contains valuable information about the nature/degree of overconstrainedness/``capacity'' (to use a modern term) of this inconsistency. From this simple ODE problem, the step to, say, first order PDEs, and the computer-assisted exploration of the compatibility of their initial conditions is an obvious one, with more natural steps to follow.

103 years ago, Jacques Hadamard published a truly influential memoir in the Princeton University Bulletin; its title was ``Sur les probl\`{e}mes aux deriv\'{e}es partielles at leur signification physique.'' \cite{hadamard1902problemes} It is worth repeating (in translation) his frontispiece quotation by Poincar\'{e} himself: ``Physics does not only give us the opportunity to solve problems --- it also makes us intuit the solution''. And it is also worth repeating the introductory paragraph of the paper: 

\begin{quote}
    The principal problems that the geometers have been led to pose relative to partial differential equations fall into two general types: the Dirichlet problem (and its analogs) in which the unknown function, defined in a certain domain, must satisfy, at every point of the boundary of this domain, a certain condition (let's consider, for fixing ideas, a second order problem); the problem of Cauchy, in which the conditions to satisfy at every point of the boundary are two in number, one giving the value of the function at that point and one giving the value of one of its first derivatives.
\end{quote} 

This three-and-a-half page memoir has cast a long shadow in the analysis (and practice) of solving partial differential equations, articulating the conditions of existence, uniqueness, and smooth dependence on (initial and) boundary conditions.

Solving partial differential equations can be thought of as a constrained manifold completion problem. Knowing the right initial/boundary conditions and the law of the PDE (its Right-Hand-Side, RHS), we can solve it, and solve it uniquely if the problem is well posed. If the problem is underconstrained, we would find (too) many solutions; if it is overconstrained, we will find none. In traditional numerical analysis, one typically does not even embark upon solving such a problem, unless one is already certain the problem is well posed (and we are not even touching upon the relation between well posedness of the full problem and well posedness of its discretization, nor on the important issue of smooth dependence on initial/boundary conditions). There have been rivers of ink (and probably Gigabytes of storage) used for publications proving, conjecturing, exploiting the well-posedness of a problem on the path towards solving it.

Enter machine learning approaches to numerically solving partial differential equations (to ``fix ideas'' as Hadamard said, let's consider the powerful approach embodied in Physics-Informed Neural Networks, PINNs \cite{raissi_physics_2017}). The approaches are mainly variational, taking advantage of automatic differentiation and the programming ease of packages like TensorFlow and Pytorch. What we will discuss is not (emphatically not!) confined to machine learning approaches to solving PDEs. Yet it is only fair to acknowledge that these methods (and the software infrastructure associated with them) naturally give rise to new ``experimental'' avenues exploring/establishing well posedness. Because of their inherently variational nature, and because of the (ridiculous for the older ones of us) ease of obtaining a solution, they allow us to do what, up to say 10 years ago, was unthinkable:

\begin{enumerate}
    \item Attempt (without fear!) to solve a problem a million times -- rather than making trebly sure it is well posed before solving it;
    \item Collect the solutions and explore (through, say, manifold learning) their statistics (their nature and intrinsic dimensionality, for example) as well as the degree to which they satisfy the loss function (the PDE residuals and the attempted initial/boundary conditions);
    \item Armed with this information, attempt to establish well posedness;
    \item If the problem is not well posed, explore what (and how much, and how many different options of how much) it would take it make it so.
\end{enumerate}

Why suddenly consider problems whose well posedness is unknown \textit{a priori}, 103 years after Hadamard? The answer is simple: because in our big data era the observations are many, and are not obtained conveniently where his ``geometers" would place them --- at the domain boundary. They may be at the boundary or the interior ; they may be observations of some arbitrary mixed derivative(s) of the function in question; they may (what riches!) be available over entire connected or disconnected subdomains in space or space-time (see Figure 1); and they may -- alas -- be intrinsically incompatible, so that the problem is both oveconstrained and underconstrained simultaneously in different parts of the solution space domain.

How to deal with these situations? More precisely: how to deal with these situations practically numerically?
This is what this paper is about: 
how to obtain numerical information that will support theoretical decisions about
\begin{enumerate}
    \item determining if the problem is well posed and, if not 
    \item what it would take to make it so. 
\end{enumerate}

Before we start, we should  discuss the elephant in the room -- the broad, powerful, useful, deep theoretical as well as applied work on data assimilation:
where does the boundary lie between data assimilation (providing an estimate of the most probable problem -- within a parametrized family of problems -- as well as the corresponding most probable solution) and what we discuss here?

There are intimate similarities, but there also differences. 
We do \emph{not} (emphatically not!) start by allowing the operator to vary; we assume we know it. 
We will choose to put the blame of failure on the prescribed initial/boundary/internal constraints -- not on the operator. 

And, when a residual cannot be satisfied within a prescribed tolerance, we are allowing ourselves to call it ``wrong'' and discuss ways to alleviate its wrongness, tracing it back to the right ``culprit'' initial/boundary/internal constraint(s), discarding it as ``corrupted'' -- and thus opening the possibility of an adversarial approach to overconstraining. 

Such situations also arise naturally in large scale randomized linear algebra. 
For linear overconstrained problems, with supernumerary prescribed initial/boundary conditions \cite{chinneck2019maximum} (see also \cite{puranik2017deletion} for nonlinear problems)  we can attempt to solve using modern randomized linear algebra implementations, like the Subspace Constrained Kaczmarz algorithm \cite{lok2023subspace}.
This is also a context in which relations between linear equation corruption,
problem overconstrainedness, and weak solution detection/characterization clearly emerges.

A final interesting connection has to do with weak solutions to PDEs \cite{evans2022partial}; overconstrained problems are known to lead, under certain circumstances, to weak solutions; and the number and nature of the encountered discontinuities may again provide numerical information towards establishing well posedness, or guiding the modifications that would help establish it.

The work is organized as follows: First, we demonstrate the basic concept of ill-posed solutions in ordinary differential equations through a simple harmonic oscillator problem, including analysis of a PINN-based solution (\cite{karniadakis2021pinns}). We then describe illustrative scenarios in solving a linear PDE, the wave equation in various ill-posed regimes. We show that we can probe not only the fact that the problem is ill posed, but also several qualitative details of \textit{how} it is so. We then show a similar analysis for a nonlinear PDE, solved only as a PINN. Finally, we consider some over- rather than under-constrained scenarios, and demonstrate how techniques from randomized iterative methods can be especially informative in these settings.

\section{Ordinary Differential Equations}
\label{sec:ODEs}
While ODEs are generically simpler to study than full-fledged PDEs, the ideas surrounding well-posedness we discuss in this work transfer naturally between the two settings. This is at a superficial and not necessarily formal level. We dedicate this section to analyzing several simple computational experiments, designed to provide insight and motivate the study of more complex differential operators in later sections.

\subsection*{Harmonic Oscillator}
We begin by considering the harmonic oscillator initial value problem:
\begin{align}
    \begin{cases}
        x''(t)=-x(t)\qc t\in[0,2\pi]\\
        x(0)=a\qc x'(0)=b.\\
    \end{cases}
    \tag{\text{IVP-1}}
\end{align}
We know that solutions of the problem take the general form:
\begin{align*}
    x(t)=b\sin(t)+a\cos(t)
\end{align*}
and in fact they form a finite (2-) dimensional vector space, since the equation itself is linear. Note that, since this problem is ``well-posed", for each pair $(a,b)$ of initial conditions, there exists \textit{a single} solution satisfying the equation.

We may now consider a similar problem where only a single initial condition is specified:

\begin{align}
    \begin{cases}
        x''(t)=-x(t)\qc t\in[0,2\pi]\\
        x(0)=a.\\
    \end{cases}
    \tag{\text{IVP-2}}
\end{align}

\begin{figure*}
		\begin{subfigure}[b]{0.5\textwidth}
			\includegraphics[width=\textwidth]{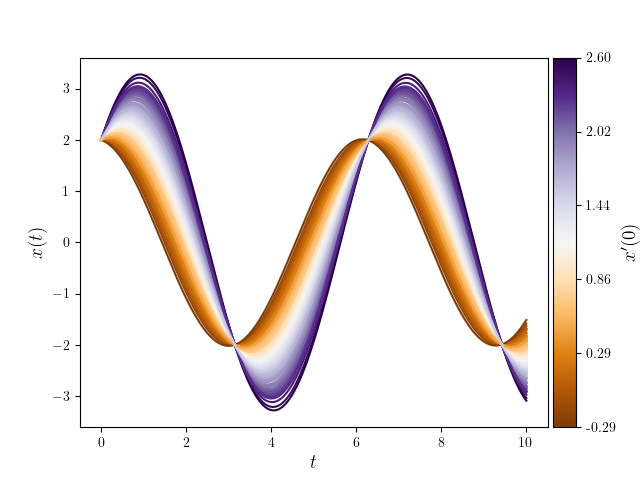}
			\caption*{IVP-2}
		\end{subfigure}
		\begin{subfigure}[b]{0.5\textwidth}
			\includegraphics[width=\textwidth]{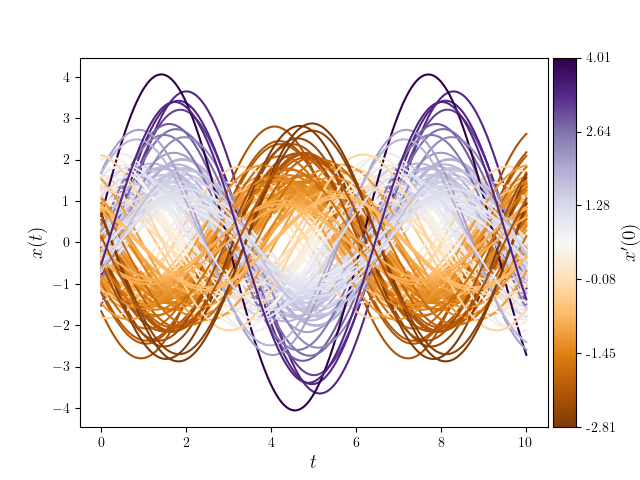}
			\caption*{IVP-3}
		\end{subfigure}
		\caption{Sampled solutions to IVP-2 with $a=2$ (left) and IVP-3 (right). Each figure depicts a set of 150 solutions, each given by separate runs of an underconstrained PINN network. Coloring represents the derivative $\hat{x}'(0)$ of the neural network with respect to time.}
            \label{fig:harmonic_oscillator_PINN}
\end{figure*}

This problem is no longer well-posed, since there no longer is a unique solution. There is now an infinite number of solutions that take the form:
\begin{align*}
    S_{\text{IVP-2}}=\qty{x(t)=c_1\sin(t)+a\cos(t):c_1\in\R}
\end{align*}
The set of solutions consists of sinusoidal waves through the point $(0,a)$ with arbitrary slope. This is, of course, a rather trivial observation stemming from the fact that we already have a general analytic solution to the harmonic oscillator. However, from a numerical perspective, we may no longer ask to `solve' IVP-2 in the traditional sense. We may instead ask to \textit{sample} some solutions from the set $S_{\text{IVP-2}}$. Conversely, from an inverse-problem perspective, we can ask what a sample of solutions to an (unknown) IVP can tell us about the IVP itself.

\subsection{Solving the Harmonic Oscillator with PINNs}

One manner of sampling from $S_{\text{IVP-2}}$ is to minimize the corresponding objective $\mcE_{NN}$ within a Physics-Informed Neural Network (PINN) framework. There, the function $x(t)$ is approximated by a smooth feed-forward neural network $\hat{x}(t)$ whose derivatives are evaluated via automatic differentiation. The objective to be minimized is given by:
\begin{align}
    \label{eqn:ho_loss_1d}
    \mcE_{NN}=\underbrace{\norm{\hat{x}(0)-x(0)}_2^2}_{\text{initial condition}}+\underbrace{\sum_{i=1}^T\norm{\hat{x}''(t_i)+\hat{x}(t_i)}_2^2,}_{\text{differential loss}}
\end{align}
evaluated on some discretization of the time domain $\qty{t_i\in[0,2\pi], i\in[T]}$.

In this case each successful \textit{single, randomly initialized} optimization run will result  in a \textit{single} solution to IVP-2, and repeating, e.g. with different network weight intializations\footnote{There is no shortage of other ways to add randomness to this process, including using stochastic optimization algorithms. Each would, presumably, provide a different bias with respect to the sampled distribution.} will yield a (probably different) sample (\cref{fig:harmonic_oscillator_PINN}, left). In a similar manner, we may even ask for PINN solutions to the differential equation without specifying \textit{any} initial condition, i.e. satisfying
\begin{align}
    x''(t)=-x(t)\qc t\in[0,2\pi]\tag{IVP-3}
\end{align}
In that case, however, the differential term of the loss is observed to very frequently result in relatively flat functions (close to a constant function under the $L^2$ norm). To obtain a more representative solution ensemble, we instead penalize such solutions by minimizing the following objective:
\begin{align}
\label{eqn:ho_loss_2d}
\mcE_{NN}=\sum_{i=1}^T\norm{\hat{x}''(t_i)+\hat{x}(t_i)}_2^2+B_\alpha\qty(\sum_{i=1}^T\norm{\hat{x}''(t_i)}_2^2),
\end{align}
where $B_\alpha$ is a compactly supported smooth bump function centered at the origin (and $\alpha$ is a parameter characterizing its width, \cref{app:ODE}).

For each IVP-2 and IVP-3 we generate 150 solutions (for $t\in[0,10]$), each resulting from a single run of an underconstrainted PINN by minimizing \cref{eqn:ho_loss_1d,eqn:ho_loss_2d} respectively. These sets of solutions are represented in \cref{fig:harmonic_oscillator_PINN}.

In these examples, we know that the theoretical solution sets are one- and two- dimensional respectively, from the analytic treatment of the IVPs. If the original operator is unknown (for example it is a black-box differential operator) the underlying dimension of a set of solutions may be unknown. In that case, using linear or non-linear dimension reduction techniques on the set of solutions we can infer, by quantifying the ``richness" of the identified solution set sample, information about the operator itself. For example, performing PCA on the set of solutions to IVP-2 reveals a one-dimensional subspace that explains the majority of the sample solutions variance, while the same method reveals a 2-dimensional subspace for solutions to IVP-3 (see \cref{app:ODE}). 
For this underconstrained problem, we are clearly interested in finding the null space of the operator - and if the operator is a linear one (e.g. a large sparse matrix, resulting from a finite difference discretization of a linear PDE), then finding its null space can be accomplished via standard methods like SVD or QR, or with modern, more scalable randomized methods such as \cite{foster2013algorithm,gotsman2008computation,park2023fast}. For general nonlinear operators, probing the dimension of the nullspace, and parametrizing it, is of course much more difficult; yet we hope that matrix-free techniques (in the spirit of matrix-free Newton-Krylov GMRES, and matrix-free Arnoldi algorithms, \cite{qiao2006,samaey2008newton}) may be useful in this direction.  

The general harmonic oscillator equation results from a second-order (linear) operator that requires two initial conditions to be solved uniquely. When the operator is nonlinear, linear dimension reduction will be inadequate to capture the intrinsic features of the space of the operator solutions, and one may need to resort to nonlinear data reduction methods, such as Diffusion Maps \cite{coifman2006diffusion} or autoencoders \cite{kramer1991nonlinear} for this purpose.

\section{Partial Differential Equations}
\label{sec:PDEs}

\subsection*{Linear PDEs (The Wave Equation)} 
\begin{figure*}[h]
    \centering
    \includegraphics[width=.18\textwidth]{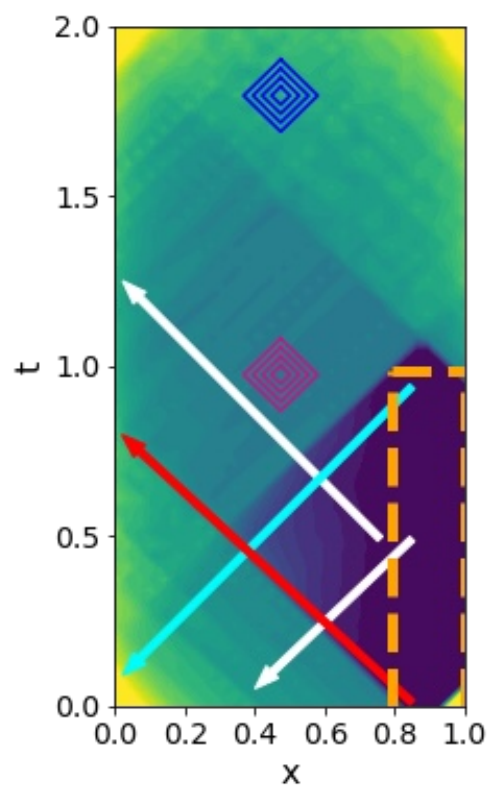}
    \includegraphics[width=.18\textwidth]{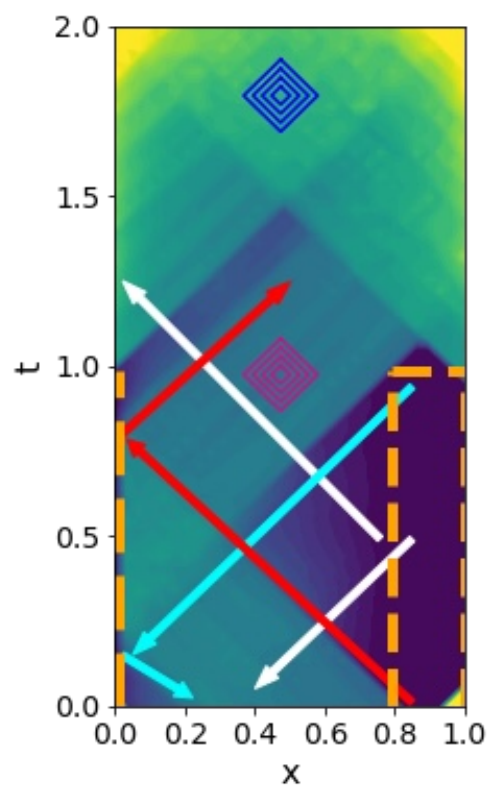}
    \includegraphics[width=0.245\textwidth]{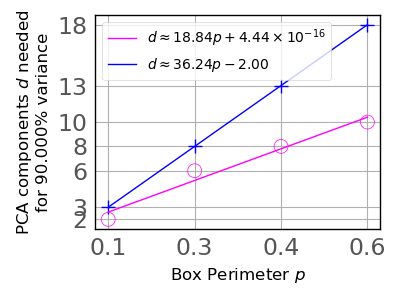}
    \includegraphics[width=0.245\textwidth]{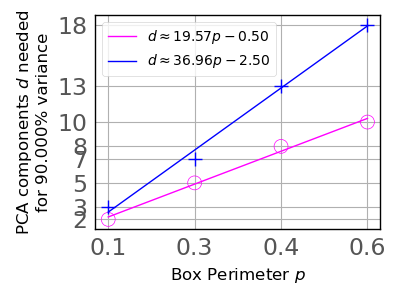}
    \caption{
        \textbf{Left:} Statistics of repeated iterative least squares minimzation solutions for (leftmost) the non-reflecting- and (second from left) the partially reflecting-BC wave equation problem. A few characteristics are included to help discuss the results. The figures also show the locations of two sets of squares (the pink and the blue ones) whose study will help estimate the solution ``richness" in the corresponding parts of the domain.  
     \textbf{Right:} (Colors of fitted lines match colors of patches in the corresponding left figures). The lines result from the solution PCA scaling with the perimeter of successive larger patches (see section \ref{sec:PDEperimeterScaling}); the relative slopes of these lines help quantify different levels of solution ``richness" prescribed by the data. The order of the latter two subfigures correspond to that of the former two: first the ``no reflector BC" case, and then the ``partial reflector BC" case.
    }
    \label{fig:iterative_WE_reflecting-stats}
\end{figure*}

We start from the solution of a well-posed problem: the wave equation in a rectangular (time $\times$ a single space dimension) domain, when a number of observations of the solution are prescribed (some of them possibly on portions of the  boundary of the space-time domain, others possibly in its interior); 
\begin{equation}
\label{eqn:law}
\frac{\partial^2u(x,t)}{\partial x^2}-c^2\frac{\partial^2 u(x,t)}{\partial t^2}=0,
\end{equation}

where $c=1$. The initial conditions are given by  $u(x,t)=0$ and $ \dot u(x,0) =  \sin(2 \pi x/c)$, while homogeneous Dirichlet boundary conditions are imposed on both spatial boundaries.
In one our case studies we only provide detailed, accurate solution data over a space-time parallelogram (see the red shaded portion of the domain in \cref{fig:iterative_WE_reflecting-stats}) and nothing else; in a second case study, in addition to these data we also prescribe a reflective boundary condition over only a portion of the left ($x=0$) time boundary.
We will pose solving each of these two problems as a linear program, arising from a regular finite difference discretization of the PDE
\begin{equation}
    \label{eqn:linProg}
    \textbf{A} \cdot \textbf{u} = \textbf{b},
\end{equation}
where $\textbf{A}$ may or may not be full rank; 
we attempt to solve  \cref{eqn:linProg} in the least squares sense. 
The formulation of the matrices $\textbf{A}$ and the vectors $\textbf{b}$ is discussed in Appendix \ref{app:PDE}. Here, $\textbf{A} \in\R^{E \times N}$ has elements $A_{m,n}$, $\textbf{u} \in \R^{N\times 1}$ has elements $\textbf{u}_{n}$, and $\textbf{b} \in \R^{E\times 1}$, where $N=n_t\cdot n_x$ (the total number of discretization points in the domain, here $n_t=60$ and $n_x=30$); $E$ is the number of linear equations we attempt to satisfy; $E$ may in general be greater than, less than, or the same as $N$.

Before we start solving this set of equations, we briefly discuss what we might expect to find
for our (clearly underdetermined) case studies. We know that the solution of the wave equation at 
some point is determined by the values propagating on two characteristics passing through the point. 
Inspecting \cref{fig:iterative_WE_reflecting-stats}(a) we can follow the effect of prescribing true solution data in the shaded space-time parallelogram by tracking the characteristics ``emanating" from this parallelogram (indicated by red or azure or white lines in the figure). In the first case study, there clearly exists a domain (shaded in deep blue) for every point of which two characteristics emanate from the known solution inside the parallelogram. We therefore expect that the solution within this blue ``piecewise pyramidal" region is unique (fixed, prescribed by the data). 
Interestingly, in our second case study, we can see that some characteristics get reflected on the (partial) left boundary condition (BC), producing additional regions in which the solution is uniquely prescribed by the data (the two new deep blue shaded regions). This discussion anticipates what we will find when we variationally solve the set of linear problem equations below; we will also discuss what happens in parts of the solution domain that are not uniquely prescribed by the data. 

\subsection{Solving the discretized Wave Equation Variationally}
\label{sec:PDEsVariationally}

\begin{figure*}[!htb]
    \centering
    \includegraphics[width=0.98\textwidth]{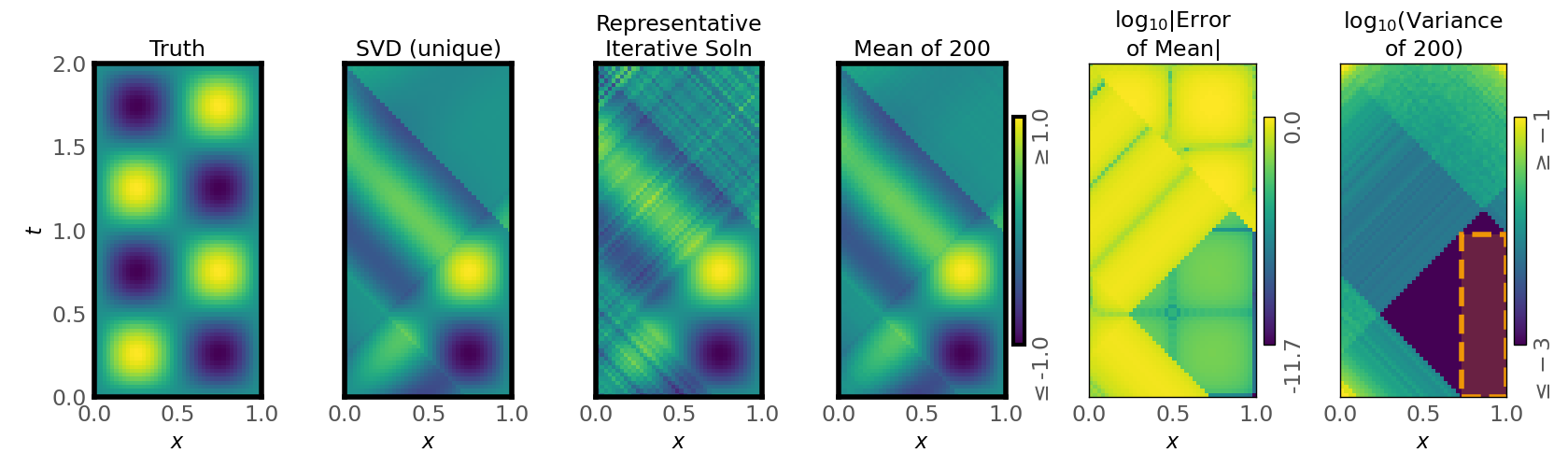}
    \includegraphics[width=0.98\textwidth]{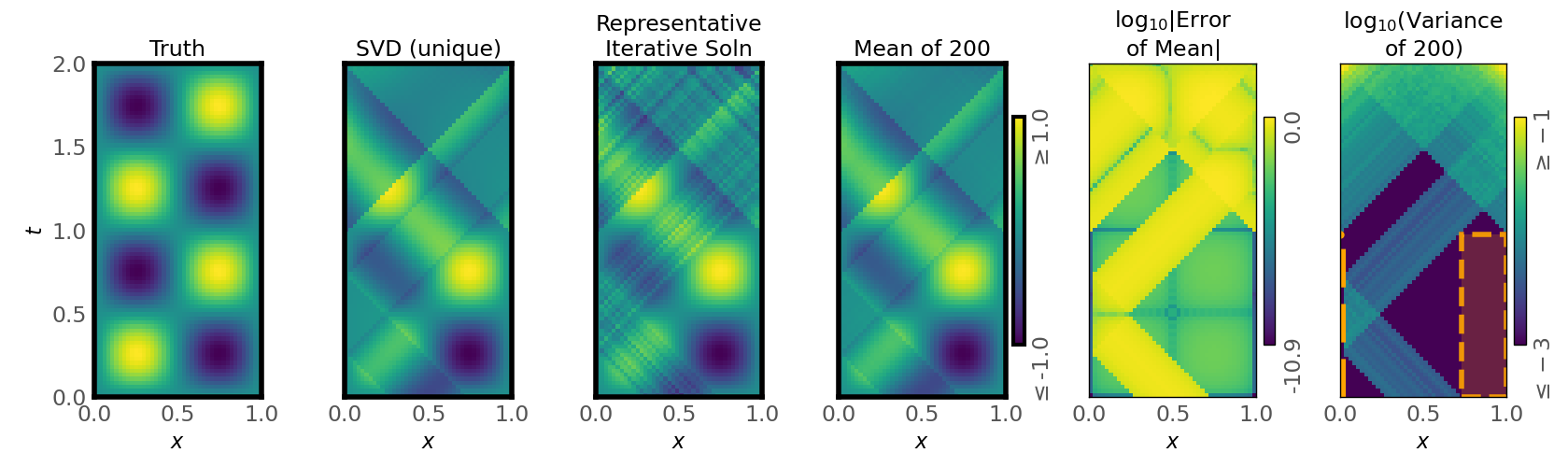}
    \caption{Truth, unique SVD solution, a single representative minimization solution, the mean of several iterative solutions, the error of this mean, and pointwise variation across the ensemble of solutions. Both non-reflecting-BC (top) and reflecting-BC (bottom) wave equation problems. The data are prescribed in the orane-bounded parallelogram. 
    \label{fig:iterative_WE_reflecting}
    }
\end{figure*}

When we use SVD to solve our (underdetermined) least-squares problem, we obtain a unique solution (cf. \cref{fig:iterative_WE_reflecting}, column 2). What happens when we use, instead, an iterative (initial guess dependent, or stochastic) minimizer to solve this problem (as we would for PINNs?).

Beginning with a true solution of the wave equation (column 1 of \cref{fig:iterative_WE_reflecting}), we prescribe the data in the red-shaded space time parallelogram (last column on the right) and solve the 
system of equations by minimizing
$$||\vb{A}\cdot \vb{u} - \vb{b}||_2^2$$
$200$ times using random seeds and an iterative minimization. Here we used the \texttt{L-BFGS-B} method from \texttt{scipy}, which implements the Broyden–Fletcher–Goldfarb–Shanno that is particularly efficient when the number of unknowns is large.
In \cref{fig:iterative_WE_reflecting}, the top row is for our case study 1 (no reflecting BC); the second row is for our case study 2 (partially reflecting BC on the left).
The single SVD solustion in each case can be seen in column 2, and 
a single sample iterative minimization solution is shown in column 3. Column 4 shows the mean of the 200 solutions, Column 5 shows the difference between the mean of 200 solutions and the first column. Most importantly, the last column 6 shows the  pointwise variation  of the set of 200 solutions in the two case studies. 
As the figure suggest, the iterative minimization recovers the correct solution in the portion of the domain colored deep blue (small variation across the random solution set) in the two rightmost columns. Clearly, those are the regions where the given solution data in the orange parallelogram prescribe {\em two} solution characteristics. In the partial reflection BC case (bottom row), two additional ``deep blue" isolated strips use the reflection of a data-prescribed characteristic to recover the correct solution for all iterative minimizations.
In the remainder of the domain the problem is not well-posed (underconstrained). 

To summarize: repeating the iterative solution multiple times allows us to probe the well-posedness of the problem by examining the variation of the solution set (the two rightmost panels of \cref{fig:iterative_WE_reflecting}) in each case study.  \cref{fig:iterative_WE_reflecting-stats} (left) discussed above is an annotated version of these two rightmost panels, with the same color shading. 

\subsection{Probing the degree of local solution richness}
\label{sec:PDEperimeterScaling}
To characterize the richness (variability) of the solutions found by randomized iterative minimizations, we define a rhomboidal set
\begin{equation}
\label{eqn:rhmobSet}
\begin{array}{rrcl}
         & t &<& x - (x^*-R) + t^*
    \\
    \cap & t &>& (x^*-R) - x + t^*
    \\
    \cap & t &>& x - (x^*+R) + t^*
    \\
    \cap & t &<& (x^*+R) - x + t^*
\end{array}\end{equation}
\newcommand{\pstar}{\vb{p}^*}
where $R=\sqrt{2r^2}$ centered at a number of points $\pstar=(x^*, t^*)$ of interest (in this case two such points, one for the pink and one for the blue rhomb sets).
We then, for a particular $\pstar$ and $r$, take the
\newcommand{\rhombAmbDim}{n_{p}}
$\rhombAmbDim$
grid locations
$(x_{i,j},t_{i,j})$
that fall within the set bounded by 
\cref{eqn:rhmobSet},
and concatenate their corresponding $\uFlattenedLetter(x_{i,j},t_{i,j})$ values into a vector featurizing the solution in the neighborhood at that center test point.
Sampling this feature vector across multiple randomly seeded minimizations
creates a cloud of ambient dimension $\rhombAmbDim$.
We then estimate the dimension of this cloud using PCA,
plot this dimension $d$ vs. the perimeter $p=8r$ of the rhomb, and subsequently fit a line to the points obtained for each $\pstar$.
The result, shown in \cref{fig:iterative_WE_reflecting-stats} (right),
is that, while the randomized solutions in well-posed regions (areas of the $x,t$ domain constrained by information from two characteristics) do not vary with the rhomb perimeter, so that the true dimension of the corresponding point clouds is always zero, the dimension of the solution set in the {\em unconstrained} (no prescribed characteristics, blue rhomb) regions grows \textit{twice as fast} with the perimeter compared to the dimension of the solution set in regions with \textit{partial prescription} (only a single data-prescribed characteristic, pink rhomb).
This is because the (discretized) perimeter allows us to (roughly) estimate the number of ``free" boundary characteristics values affecting the richness of possible solutions in the rhomb. 
Analyzing the statistics of the randomized solution set thus yields an estimate of the local number of ``missing constraints" (unspecified characteristics).

\subsection{Solving -- and probing well-posedness -- via PINNs}
\label{sec:PINNs}

Briefly, a trained PINNs is a neural network-parametrized function $\pinnNet(x,t)$ that minimizes a two-term loss.
In the first, self-supervised term of the loss, we require that, when $\pinnNet$ is substituted for $u$, \cref{eqn:law} is approximately satisfied
in a mean square sense.
Crucially, the partial derivatives of $\pinnNet$ in the law expression are evaluated with automatic differentiation.
In a second, supervised loss term, we require that $\pinnNet(x,t)\approx u_\mathrm{GT}(x,t)$: the PINN reproduces the known (Ground Truth) values prescribed at selected points.

\subsubsection{WE with PINNs}
\label{sec:WELinnearPDEPinn}

We solve the linear wave equation problem described in this section as a PINN \cite{Raissi2017b}.

For each of our two case studies, the network $\pinnNet$ has 8 hidden layers of width 32 with $\tanh$ activation functions.
Batches consist of 363 supervised (drawn randomly from a grid of experiment-dependent prescribed ground truth points) and 363 self-supervised points (drawn randomly across the solution domain of interest).
Supervisory data is obtained by solving a regular finite difference discretization (32 steps wide in $x$ and 65 tall in $t$) well-posed problem with the SVD method.
PINN training uses Adam in TensorFlow \cite{tensorflow2015-whitepaper} with a learning rate of $10^{-4}$.

A few resulting PINN solutions and, importantly, the  statistics across multiple runs are shown in \cref{fig:WE_Pinn}.  They corroborate the results obtained through finite difference iterative minimization solutions: the portions of the domain where the solution is uniquely prescribed (resp. partially prescribed) by the data become clearly visible. 

\subsubsection{Nonlinear PDEs (The KS Equation)}

\textbf{KSE PINN in a T-shaped space-time domain}

We now turn to a nonlinear parabolic PDE, the Kuramoto-Siavshinsky equation \cite{kevrekidis1990saddle} in one space dimension with periodic boundary conditions, solved with a tanh-activation PINN with a few thousand network parameters.
Our PDE law is reads now
\begin{gather}
    \label{eqn:KSE}
        \frac{\partial u(x,s)}{\partial s} = f(u(x,s)) \equiv \nonumber \\
    -4 \frac{\partial^4 u(x,s)}{\partial x^4}
    -\alpha
    \left(
        \frac{\partial^2u(x,s)}{\partial x^2} + u(x,s) \frac{\partial u(x,s)}{\partial x}
    \right)
\end{gather}
with the parameter value  $\alpha=53.3$ and for $x\in[0,2\pi)$.

\begin{figure*}[h]
    \centering
    \includegraphics[width=0.8\textwidth]{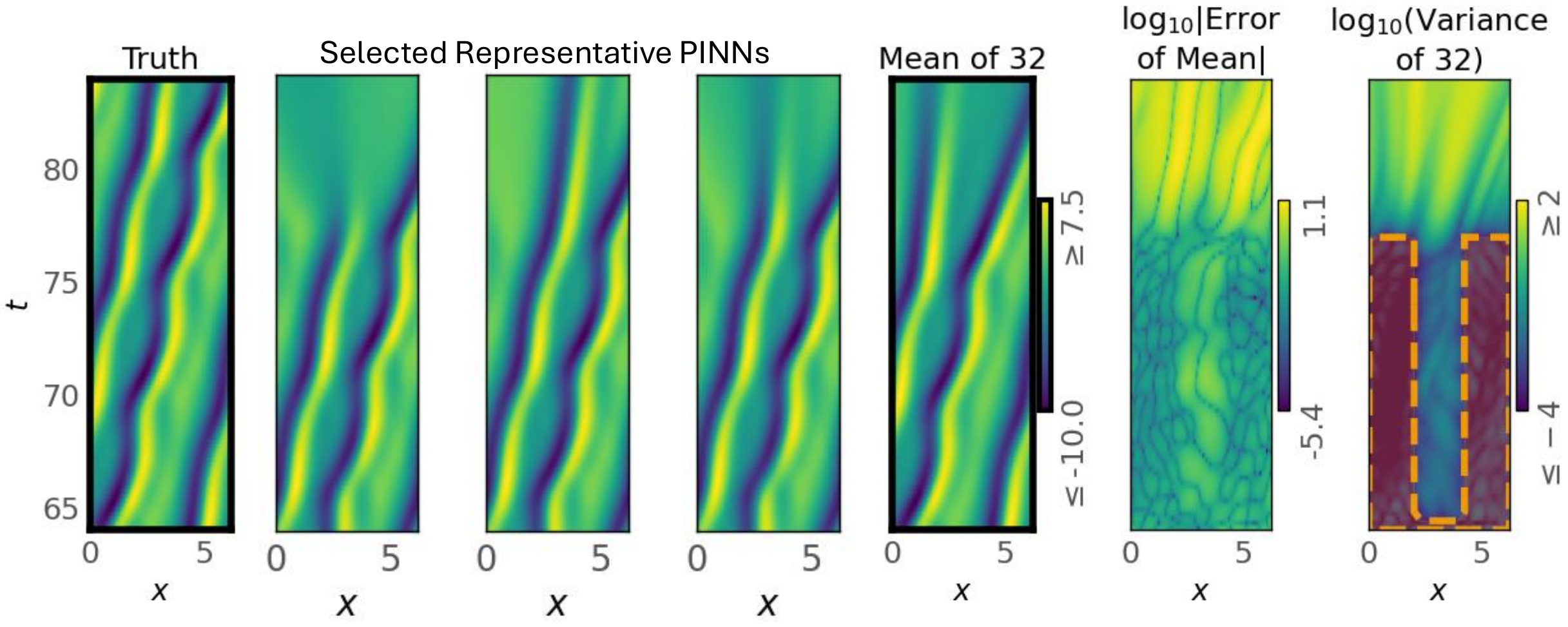}
    \caption{
        Statistics of some randomized PINN solutions of the space-time T-shaped domain for KSE for the KSE problem.
         From left to right: the true solution, three samples of randomized solutions, the mean of the randomized solution set, the error of the mean of the set and its variance (with the two-column domain where the data are prescribed outlined in orange). The complement of the supervisory domain is the T-shaped space time solution domain. } 
    \label{fig:pinnN-up}
\end{figure*}

Ground truth supervisory data is generated by a pseudospectral semidiscretization of the PDE (using \texttt{scipy.fftpack}\cite{2020SciPy-NMeth}) to formulate a high-dimensional ODE. This is in turn integrated with \texttt{scipy}'s BDF \cite{Byrne1975}, using an absolute tolerance of $10^{-5}$ and a relative tolerance of $10^{-7}$.
A sinusoidal initial condition was used, which (after an initial ``burn-in" transient attracted to the inertial manifold of the equation) approached the {\em modulated-traveling-wave} long term  seen in the leftmost panels of \cref{fig:pinnN-up}.
Minibatches consist of 1,000 $x,t$ self-supervised points sampled uniformly from the space-time domain (not from a grid),
and 8,892 supervised points (to be discussed) that are repeated for every minibatch.
We use 5 hidden layers of width 62, and the Adagrad optimizer with a learning rate of $5\times 10^{-2}$.
Minibatched training is followed by 64 iterations of conjugate gradient
in \texttt{scipy.optimize.minimize}
with gradients provided by TensorFlow.

Staring from this known solution (with periodic BC) we formulate a new initial/boundary value problem, set up in the rightmost panel of \cref{fig:pinnN-up}: we {\em prescribe} the solution in the space-time domain outlined in orange in the panel, and now try to solve the PDE in the complementary T-shaped space-time domain. It stands to reason that the prescribed space-time data are sufficient to define a well-posed problem in the "stem" of the T-shaped domain (and we already know what the unique solution will be there). But in the top ``bar" of the T-shaped domain, where we now have no boundary conditions in space, the problem cannot be well posed. 

We again repeatedly solve this partially self-supervised (PDE law), partially supervised (prescribed solution within the orange bounded subdomain) problem. 
At each iteration, we minimize a sum over a static supervised batch and a randomly-generated self-supervised (law) batch.
We repeat the training  a handful ($M\approx20$) of times, and evaluate some of our local-statistic metrics (same as in the Wave Equation case).  
A few sample solutions can be seen in the second to fourth columns of \cref{fig:pinnN-up}: clearly, the {\em same (visually)} solution has been obtained in the ``stem" of the T, and {\em different} solutions have been obtained every time at the ``top bar" of the T (the underconstrained region). 
The last column in the figure shows the variance of the solution ensemble in the T-shaped domain: it is visually clear that, practically, the same solution is obtained at every attempt with the central corridor (the stem of the T) and different solutions are obtained at every attempt in the horizontal ``bar", where no BCs have been imposed and the problem is underdetermined.

\section{The overconstrained case: Ordinary Differential Equations}

In previous sections, we have considered only \textit{feasible} problems that have at least one solution satisfying \textit{both} the law (an ODE or a PDE) \textit{and} the prescibed boundary data. For {\em overconstrained problems} we face a complementary challenge: there may be \textit{no} solution that satisfies the law {\em and the prescribed data} in a classical sense. 

Such a case, for example, would manifest as a linear problem obtained through a finite difference scheme with too many linear constraints, having no solutions.
Whatever the source of the supernumerary inconsistent constraints (noisy observations, adversarial ``attacks", or inadvertent mixing observations
from different realizations), the main premise is that the resulting linear problem has \textit{meaningful consistent sub-problems}. The proposed approach to identify them is to consider {\em subsets of contradictory boundary conditions} as ``adversarial'' data that should be identified and disregarded. 

We illustrate this approach on one of the simplest possible problems, the ODE
$y'' = 0$
that is is solved by the linear functions
$$y(t) = y'(0) t + y(0) $$
and requires exactly two (non-redundant) constraints for uniqueness (e.g. $y'(0), y(0)$). We consider a straightforward way of numerically solving this problem by discretizing it in time by $n$ inner equidistant grid points,  approximating $y''(t)$ with a 3-point stencil, and solving the least squares problem $\vb{A} \vb{y} = \vb{b}$. Here, the corresponding matrix $\vb{A} \in \R^{n + 2 \times n+2}$ has a tri-diagonal top $n \times (n+2)$ part and two indicator rows ($n+1$ and $n+2$), and $\vb{b} \in \R^{n+2}$ has its first $n$ coordinates zero.

In the case three or more (possibly contradictory) boundary conditions are provided for $y''(t) = 0$, the corresponding linear problem $\vb{A} \vb{y} = \vb{b}$  might not have an exact solution, if no line $y(t)$ satisfies all given constraints.

Then, we can find a number of solutions {\em partially} satisfying the given information by various relaxations such as: 
\begin{enumerate}
    \item By varying the  weighting between the rows in the least squares problem; this corresponds to emphasizing the importance of satisfying {\em either} the law {\em or} the constraints preferentially;
    \item By finding individual equations (rows in $\vb{A}$)  {\em to drop until the resulting system becomes consistent}. This approach can ensure (a) that the resulting system has an exact solution and (b) that the law is still everywhere satisfied. This then becomes a discrete optimization problem: finding a subset of equations to drop.
\end{enumerate}

\cref{fig:Ayb_oc_solns} below illustrates a simple case when there is exactly {\em one} constraint too many, and an exhaustive search can be performed.

\begin{figure*}[!htb]
    \centering
    \includegraphics[width=0.27\textwidth]{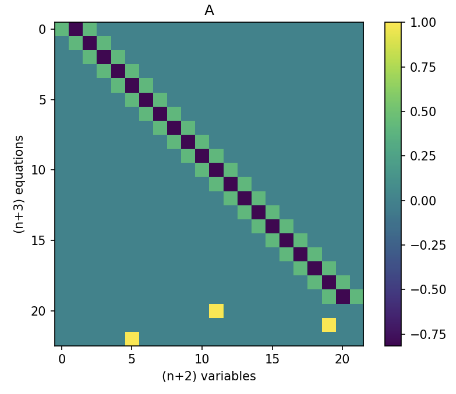}\includegraphics[width=0.7\textwidth]{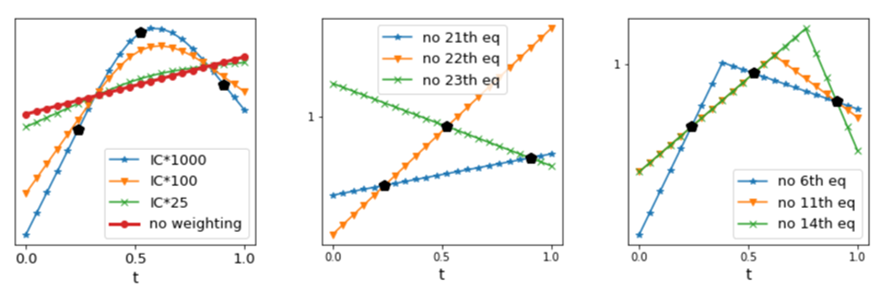}
    \caption{Left: heatmap of $\mathbf{A}$ with a tri-diagonal top part --corresponding to the ODE law ($y''(t)=0$),-- and indicator rows in the last $3$ positions, corresponding to (contradictory) boundary conditions; Left-middle: Multiplying the last $3$ equations in $\vb{A} \vb{x} = \vb{b}$ by a large factor forces the least squares solutions to this system to violate the ODE law more and approximate the prescribed conditions  better; Right-middle: ``leaving out" one of the last $3$ equations leads to a consistent solution with only $2$ boundary conditions; Right: ``leaving out" one of the equations from the top operator block of $\mathbf{A}$ leads to discontinuities: solutions satisfying the equation in portions only of the domain (related to weak solutions).}
    \label{fig:Ayb_oc_solns}
\end{figure*}

\begin{figure}[!htb]
    \centering
    \includegraphics[width=0.47\textwidth]{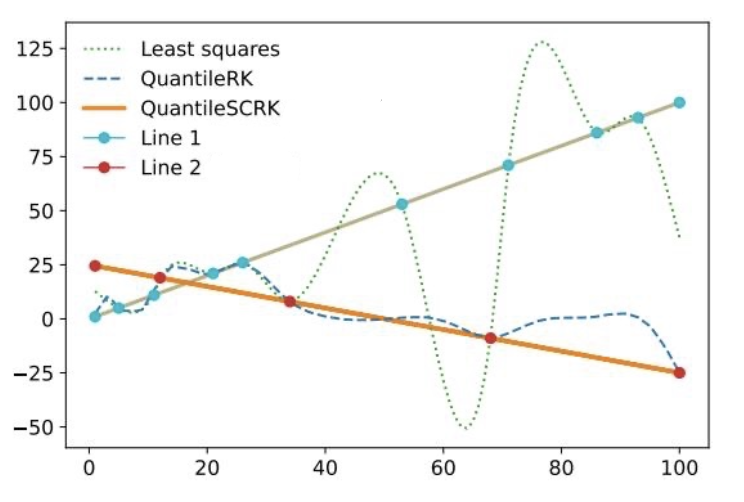}
    \caption{An example with $15$ contradictory boundary conditions ($10$ consistent blue and $5$ consistent red ones); these can be thought of as coming from inadvertently combining data from two different solutions for the $\ddot y = 0$ problem, shown by Line 1 (orange) and Line 2 (gray). The least squares solution interpolates between all boundary conditions and is not informative. QuantileRK (which here struggles to converge) and QuantileSCRK attempt to drop certain amount of ``adversarial" equations on the fly, using the statistics of the residuals during the iteration process. QuantileSCRK additionally uses the knowledge of which equations correspond to the ODE law and never violates them; this allows it to find both possible solution lines (depending on the initial seed and inner randomness of the method) efficiently.}
    \label{fig:Ayb_oc_solns_multiple_drops}
\end{figure}

In the general case, when more than one point must be “left out” to obtain a consistent system, the exhaustive search scales exponentially with $n$ and we face an instance of a generally NP-hard maximal feasible subsystem (MAXFS) problem to find which equations (constraints) to keep so that a unique solution for the linear system is obtained. The MAXFS problem can be approached by methods like random sample consensus (RANSAC \cite{fischler1981random}), see also  \cite{puranik2017deletion,chinneck2019maximum, amaldi2005randomized}. However, because of the special structure of the problem (e.g., it makes sense to assume that we know which equations in $\vb{A}$ describe the law, so that we do not try to ``leave them out") more scalable randomized iterative methods can be employed. For example, the recently introduced Quantile Subspace Constrained Randomized Kaczmarz algorithm \cite{lok2023subspace}[Algorithm 2] (based on \cite{haddock2022quantile} and \cite{strohmer2009randomized,kaczmarzoriginal}) converges to solutions that satisfy large fractions of the total number of equations at an exponential expected rate (see \cref{fig:Ayb_oc_solns_multiple_drops}).

\section{Discussion}
The purpose of this paper is to illustrate that one can use data-driven approaches, powered by the facility in solving PDEs via ML-assisted computations, to probe mathematical questions about the nature of PDE solutions - and in particular, to {\em explore and numerically quantify the well-posedness of solutions of ODEs and PDEs}. 
Solving a discretized problem variationally, through an iterative optimizer, many times, and analyzing the statistics of the solution set, enables us to characterize underconstrained problems. At the same time, randomized linear algebra techniques allow us to characterize overconstrained problems. 
The ease in solving problems variationally many times with different random seeds provides a dataset of (successful or failed) solutions. Data analysis of these solution sets guides our discovery of obstructions to well-posedenss. It can even suggest which the problematic constraints are that we could/should omit in order to obtain a unique solution. 

Given the discrete nature of ML or any other computational algorithms, our methods probe the underlying solution at discrete set of points in the space-time domain, necessarily implying an approximation of the underlying solutions in the general case. However, if the solution has some specific structural properties such as analyticity, then specifying the solution over patches of space-time suffices to extend it uniquely to the whole domain. In a similar spirit, for special classes of solutions, such as those lying in band-limited spaces (with compactly supported Fourier transforms), probing the solution at sufficient number of discrete space-time points is sufficient to uniquely determine the solution due to the Shannon-Nyquist theorem. Otherwise, our statements are approximate, and additional estimates to quantify the discretized residual to the underlying continuous residual in terms of quadrature errors become necessary.  

Future work includes developing randomized numerical linear algebra and stochastic optimization techniques that learn to satisfy parts of the system corresponding to non-contradictory initial/boundary conditions or ODE and PDEs. The nature of differential equation law gives us domain-specific information about the linear equations of the corresponding discretization that can or cannot be satisfied together (such as, through the structure of characteristics in the wave equation). This information is not present in general purpose linear solvers and optimization techniques, thus motivating the development of problem-specialized methods. For example, the row-action methods such as Randomized Kaczmarz allow to impose separate conditions on the equations of the system and this is how the Subspace Constrained version ensures that the underlying operator is not allowed to vary \cite{lok2023subspace}. 

Then, further work includes developing a general pipeline for testing the dimensionality of solution spaces in various parts of the domain resulting in practical suggestions on the additional data needed to ensure well-posedness of a given problem.

\section{Acknowledgments}
The work of TB and IGK was partially supported by the US Department if Energy and by the US National Science Foundation. The work of GAK was performed while in graduate school at UMass Amherst in 2020. ER was partially supported by NSF DMS-2309685 and also thanks Jackie Lok for the help with the plots.

\bibliography{main}
\bibliographystyle{ieeetr}

\onecolumn

\appendix

\section{Underconstrained ODEs}
\label{app:ODE}
This section provides some additional information regarding the numerical computations of \cref{sec:ODEs}.

The bump function $B_\alpha$ used to penalize near-constant solutions is defined as 
\begin{equation}
        B_\alpha(x)=
        \begin{cases}
        \exp[1/(x^2-\alpha)]\qc &\abs{x}<\sqrt{\alpha}\\    
        0\qc &\abs{x}\geq\sqrt{\alpha}.
        \end{cases}
\end{equation}
Here $\alpha$ is a positive scalar parameter that controls the support of the function. In particular, we use $\alpha=0.4$ for the results included in \cref{fig:harmonic_oscillator_PINN}. 

Performing a PCA decomposition of the solutions produced by PINNs yields, for the leading three principal components, relative weights of 2.1e+2, 1.4e-3, and 4.0e-4 for IVP-2, and 1.1e+3, 3.8e+2, and 2.7e-4 for IVP-3. These indicate the existence of a one-dimensional (resp. two-dimensional) linear subspace(s) that faithfully represent the solution set(s) for each IVP respectively. The computation was performed using the built-in library of \texttt{sklearn} in \texttt{python} \cite{scikit-learn}.

\subsection*{PINN solutions of IVPs}
For each of  (IVP-2) and (IVP-3) we generate a total of 150 solutions using PINN architectures. We use a uniform discretization of the time interval with 1000 points for $t\in[0,10]$ on which the network loss is evaluated. All networks consisted of 5 fully-connected linear layers with a width of 20, and $\tanh$ or $\sin$ activation functions applied to the first four layers. The architectures are implemented in \texttt{pytorch} \cite{paszke2017automatic} and optimized using Adam \cite{kingma2014adam}.

\section{The Matrix \textbf{A} and the vector  \textbf{b} for the Wave Equation formulations.}
\label{app:PDE}

This section details the construction of the matrix \(\textbf{A}\) and the vector \(\textbf{b}\) from \cref{eqn:linProg}, which arise in the discretization of the wave equation. The spatial domain \([a, b]\) is discretized using \( n_x =30\) equidistant points, and the temporal interval \([0, T]\) is divided into \( n_t =60\) equal steps:

\begin{equation}
    \Delta x = \frac{b - a}{n_x - 1}, \quad \Delta t = \frac{T}{n_t - 1}.
\end{equation}

For numerical stability, the Courant-Friedrichs-Lewy (CFL) condition must be satisfied:

\begin{equation}
    \frac{c \Delta t}{\Delta x} \leq 1,
\end{equation}
\noindent with $c=1$.

Using the Central Time-Central Space (CTCS) scheme, the finite difference approximation at each interior point \((n, i)\) is:

\begin{equation}
    u^{n+1}_i = 2u^n_i - u^{n-1}_i + C^2 (u^n_{i+1} - 2u^n_i + u^n_{i-1}),
\end{equation}

where \( C^2 = (c \Delta t/\Delta x)^2 \). This leads to a sparse (banded) linear system.

\subsection{Structure of Matrices \textbf{A} and \textbf{b} in the ``standard" well posed case}
\label{sec:struc}
We first discuss the structure of $\textbf{A}$ and $\textbf{b}$ that typically results from the implementation of a well-posed finite difference approximation. The initial conditions are given by  $u(x,t)=0$ and $ \dot u(x,0) =  \sin(2 \pi x/c)$, while homogeneous Dirichlet boundary conditions are imposed on both spatial boundaries.

This is not one of the (underconstrained) case studies presented in Section \ref{sec:PDEs}, yet it is included for the sake of completeness, since it leads to the unique solution presented in Figure \ref{fig:iterative_WE_reflecting} (top/bottom leftmost plot). \textit{Parts of this solution} (i.e. the areas bounded by a dashed orange line in the rightmost part of Figure \ref{fig:iterative_WE_reflecting}) constitute the data prescribed, subsequently, in the two case-studies discussed in Section \ref{sec:PDEs}.

The rows, $r$, of matrix \textbf{A} encode three types of equations:

\begin{itemize}
    \item \textbf{Initial Condition Rows:} These enforce the initial condition for $\textbf{U}$ at \( t=0 \):
    \begin{equation}
    \begin{aligned}
        % A_{r, 0, i} &= 1.0, &
        A_{r,i+n_x} &= 1.0, &
        b_r &= u_0(i \cdot \Delta x).
    \end{aligned}
    \label{eqn:IC1}
    \end{equation}

     Here $1\leq i \leq n_x$, populating the first $n_x$ rows of the matrix $\textbf{A}$
    
    The initial condition for the first derivative, $\dot{\textbf{U}}$ leads to:

    \begin{equation}
    \begin{aligned}
        % A_{r, 1, i}   &=  1.0, &
        A_{r, i+  n_x}   &=  1.0, &
        % A_{r, 0, i}   &= -1.0 - C^2,      \\
        A_{r, i+n_x}   &= -1.0 - C^2,      \\
        % A_{r, 0, i-1} &=  \frac{1}{2} C^2, &
        A_{r, i-1+n_x} &=  \frac{1}{2} C^2, &
        % A_{r, 0, i+1} &=  \frac{1}{2} C^2,  \\  
        A_{r, i+1+n_x} &=  \frac{1}{2} C^2, \\
        b_r           &=  u_0t(i \cdot \Delta x) \cdot \Delta t,
    \end{aligned}
    \label{eqn:IC2}
    \end{equation}

     Here $2\leq i \leq n_x-1$ populating another $n_x-2$ rows of the matrix $\textbf{A}$.

    \item \textbf{Boundary Condition Rows:} These enforce the boundary conditions at \( x = a \) (left boundary) and \( x = b \) (right boundary):

    \begin{equation}
    \begin{aligned}
        % A_{r, n, 0} &= 1.0, &
        A_{r, n \cdot n_x} &= 1.0, &
        b_r &= 0.0, \\
        % A_{r, n, n_x-1} &= 1.0, &
        A_{r, (n+1) \cdot n_x -1} &= 1.0, &
        b_r &= 0.0,
    \end{aligned}
    \label{eqn:boundary}
    \end{equation}

     where $1<n<n_t$, populating $2\times(n_t-1)$ additional rows of the matrix $\textbf{A}$.
    
    \item \textbf{Wave Equation Rows:} These encode the CTCS finite difference approximation of the equation for the interior points:

    \begin{equation}
        \begin{aligned}
        % A_{r, n+1, i} &= 1.0, &
        % A_{r, n, i} &= -2.0 + 2.0 C^2, \\
        % A_{r, n-1, i} &= 1.0, &
        % A_{r, n, i-1} &= -C^2, \\
        % A_{r, n, i+1} &= -C^2 &  b_r &= 0.0.
        A_{r, i+(n+1)\cdot n_x} &= 1.0, &
        A_{r, i+n \cdot n_x} &= -2.0 + 2.0 C^2, \\
        A_{r, i+(n-1)\cdot n_x} &= 1.0, &
        A_{r, i-1+ n \cdot n_x} &= -C^2, \\
        A_{r, i+1 + n \cdot n_x} &= -C^2 &  b_r &= 0.0.
        \end{aligned}
        \label{eqn:lawcoord}
    \end{equation}

    Here, $1<n<n_t$ and $1<i<n_x$ populating $(n_x-2)\times(n_t-2)$ additional rows of the matrix $\textbf{A}$, for a total of $N=n_t\times n_x$ rows; for the well posed problem this matrix is full rank.
\end{itemize}

\begin{figure}[h]
    \centering
    \includegraphics[width=0.6\linewidth]{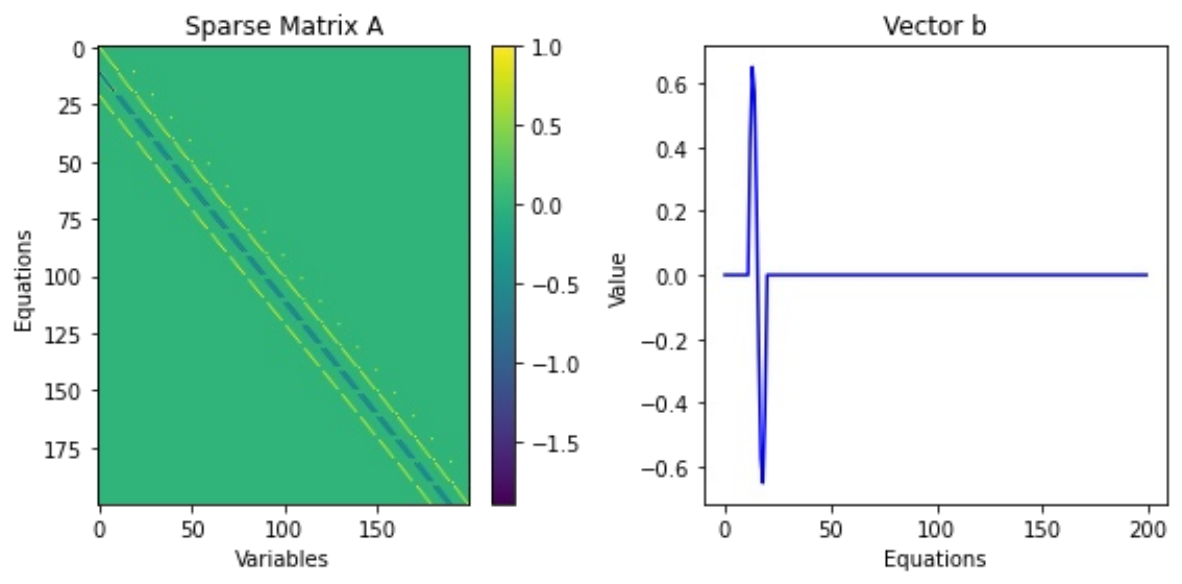}
    \caption{The sparsity pattern of $\textbf{A}$ and the values of $\textbf{b}$ for (relatively small, smaller than those used in the actual computations) values of $n_x=10$ and $n_t=20$ (solely for visualization purposes). 
The rows of $\textbf{A}$ and the elements of $\textbf{b}$ are populated based on equations  \ref{eqn:IC1},\ref{eqn:IC2}, \ref{eqn:boundary} and \ref{eqn:lawcoord}.}. 
\label{fig:enter-label}
\end{figure}

The well-posed system $\textbf{A} \cdot \textbf{u}=\textbf{b}$ gives rise to a unique solution, shown in Figure \ref{fig:iterative_WE_reflecting} (first plot on the left). 

As mentioned above, we use this solution, or rather, the part of it contained in the orange rectangle shown in Figure \ref{fig:iterative_WE_reflecting} (rightmost plot), to modify the matrix $\textbf{A}$ and the vector $\textbf{b}$, in conducting the experiments detailed in the two case-studies of Section \ref{sec:PDEs}. We now describe these modifications.

 \subsection{The case-study without a partially reflecting boundary}
 \label{app:withRef}
 The rows of this version of the matrix $\textbf{A}$, are populated as follows:
 \begin{itemize}
     \item the solution (from the well-posed problem, $u_{WP}$) within the orange parallelogram (cf. Figure \ref{fig:iterative_WE_reflecting} - plot in $6^{th}$ column):
      \begin{equation}
         \begin{aligned}
        % A_{r, 1, i}   &=  1.0, \\
        A_{r, i+n_x}   &=  1.0, \\
        b_r           &=  u^i_{WP}
    \end{aligned}
    \label{eqn:super}
    \end{equation}
     \item the law equation as described in Section \ref{sec:struc}. 

In this version of A (a) there are no rows corresponding to Initial/Boundary conditions; (b) there exist rows corresponding to prescribing the solution of the equation within the orange parallelogram.

 \end{itemize}

\subsection{The case-study with partially reflecting left boundary}
In the case with partially reflecting left boundary, $\textbf{A}$ and $\textbf{b}$ are formulated as in the case without reflection (detailed in Section \ref{app:withRef}), but now with the addition of $n_t/2$ rows corresponding to the boundary condition applied along the orange dashed line marking the reflecting portion of the boundary, on the left side of the rightmost plot in Figure \ref{fig:iterative_WE_reflecting}:
\begin{equation}
    \begin{aligned}
        % A_{r, n, 0} &= 1.0, &
        A_{r, n \cdot n_x} &= 1.0, &
        b_r &= 0.0, \\
    \end{aligned}
    \label{eqn:boundary1}
    \end{equation}

\begin{figure}[h]
    \centering
    % First subfigure
    \begin{subfigure}[b]{0.6\textwidth}
        \centering
        \includegraphics[width=\textwidth]{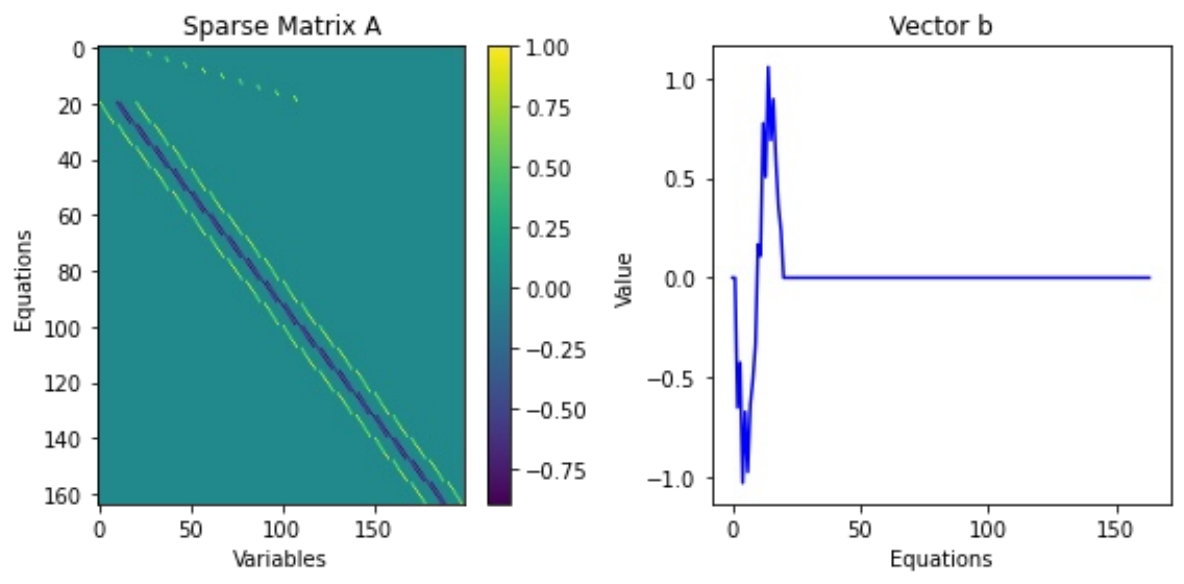} 
        
        \label{fig:sub1}
    \end{subfigure}
    \hfill
    % Second subfigure
    \begin{subfigure}[b]{0.6\textwidth}
        \centering
        \includegraphics[width=\textwidth]{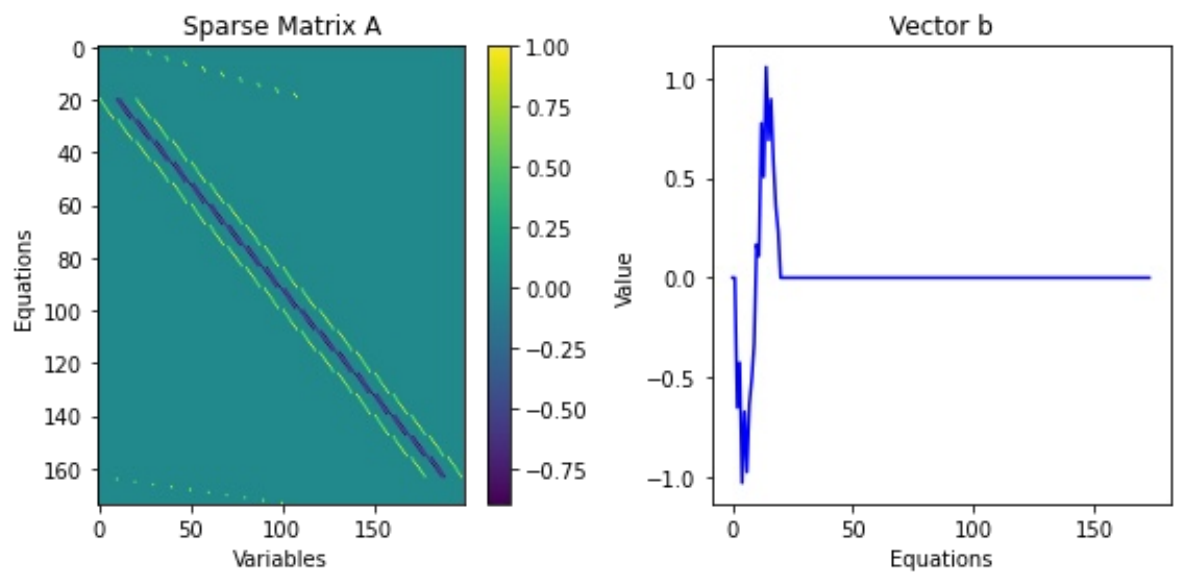}
        
        \label{fig:sub2}
    \end{subfigure}
    \caption{Visualization of the matrix $\textbf{A}$ and the vector $\textbf{b}$ in the case without reflection (top) and with partially reflecting left boundary (bottom). In both cases, the first $10$ rows of the matrix $\textbf{A}$ are populated by contributions of the prescribed solution within the parallelogram. The next $(n_x-2)\cdot(n_t-2)$ are populated by the law equation. In the case with partially reflecting boundary (bottom) $n_t/2$ additional rows are appended, implemeting this BC. \textit{Note: Solely for visualization purposes, here $n_x=10$ and $n_t=20$ are much smaller than those used for the actual computations in the paper.}}
    \label{fig:Ab_with_no_ref}
\end{figure}

A visualization of $\textbf{A}$ and $\textbf{b}$ in the cases (a)  with and (b)  without reflection is shown in Figure \ref{fig:Ab_with_no_ref}. We use SVD to derive a unique solution of these \textit{underdetermined} systems, resulting in the second column entry (top for the case without reflection and bottom for the case with reflection) of Figure \ref{fig:iterative_WE_reflecting}. The same systems are then solved variationally, leading to different solutions (when initialized with different random seeds) as discussed in Section \ref{sec:PDEsVariationally}. A single representative such solution is included in the third column of Figure \ref{fig:iterative_WE_reflecting} (top and bottom for no reflection/reflection, respectively).

\section{Wave Equation PDE Solutions via PINNs}

Having probed the well-posedness of two one-dimensional wave equation problems by solving the set of linear discretization equations via least squares minimization, we repeat the exercise solving the same two problems via PINNs. The figure below contains results analogous to those in \cref{fig:iterative_WE_reflecting} in the main text. Each row contains results for one of the two formulations: no left boundary condition (above) and a reflector boundary condition on part of the left boundary (below).
The nominal solution (``truth") is the first column. Three representative minimization PINN solutions follow, as well as the mean of 28 (resp. 32) solutions from random seeds. The error of the mean is the next to last column. The final (and most important) colum shows the variance of 28 (resp 32) random seed solutions in the two cases. The prescribed data and the "extra" BC are denoted in broken orange lines. The portions of the domain where the problem appears well posed are precisely the dark (low) variance regions, where the same solution is practically recovered for every random seed. The rationalization of these ``well-posed" domain shapes is the same as above. 

\begin{figure*}[!htb]
    \centering
    \includegraphics[width=0.98\textwidth]{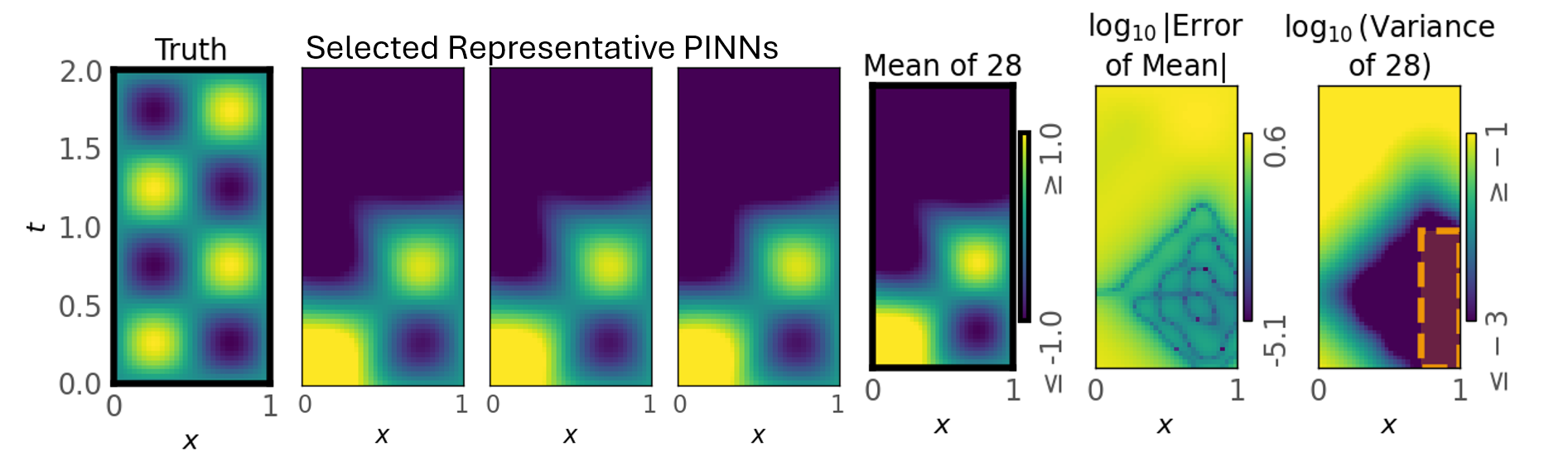}
    \includegraphics[width=0.98\textwidth]{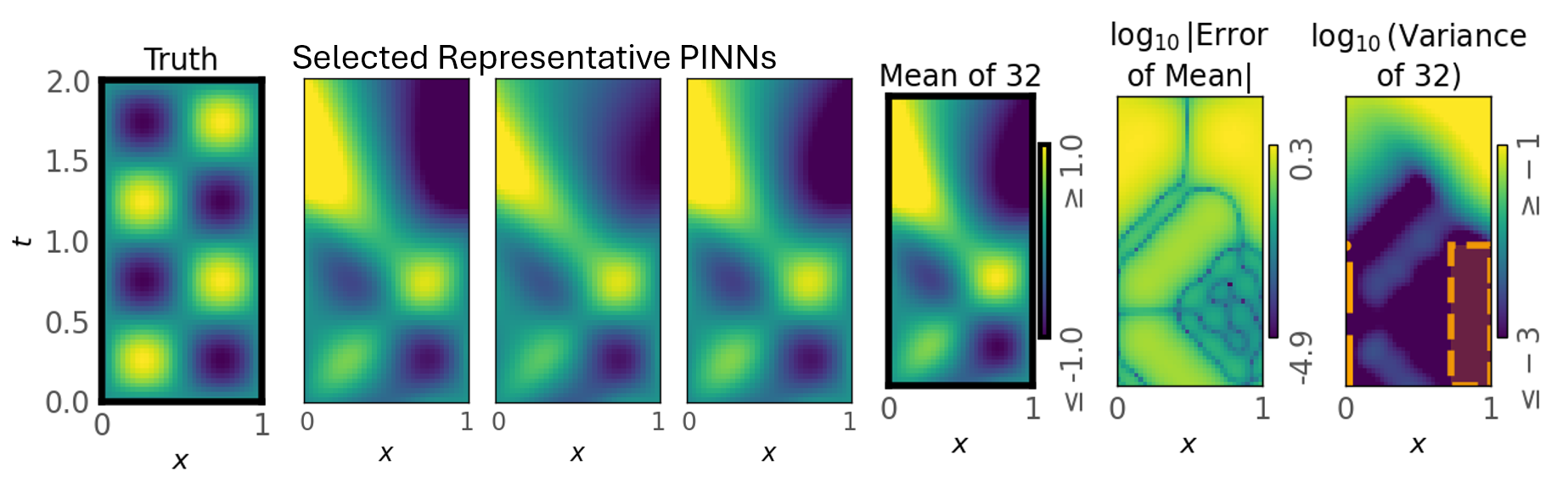}
    \caption{Truth, representative minimization solutions, the mean of several iterative solutions, the error of this mean, and pointwise variation across the ensemble of solutions. Both non-reflecting-BC (top) and reflecting-BC (bottom) wave equation problems. The data are prescribed in the orange-bounded parallelogram. 
    \label{fig:WE_Pinn}
    }
\end{figure*}

\end{document}